\documentclass[review]{elsarticle}
\usepackage{hyperref}
\usepackage{amsmath}
\usepackage{enumitem}
\usepackage{makecell}
\usepackage{booktabs}
\usepackage{multirow}
\usepackage{amssymb}
\usepackage{algpseudocode}
\usepackage{algorithm}
\usepackage{subfigure}
\usepackage{url}
\usepackage{color}

\newtheorem{assumption}{\bf Assumption}

\begin{document}

\begin{frontmatter}

\title{Data-Free Quantization via Mixed-Precision Compensation without Fine-Tuning}

\author[1]{Jun Chen\fnref{first}}
\ead{junc@zju.edu.cn}
\author[1]{Shipeng Bai\fnref{first}}
\ead{shipengbai@zju.edu.cn}
\author[1]{Tianxin Huang}
\ead{21725129@zju.edu.cn}
\author[1]{Mengmeng Wang}
\ead{mengmengwang@zju.edu.cn}
\author{Guanzhong Tian\fnref{address1}\corref{mycorrespondingauthor}}
\ead{gztian@zju.edu.cn}
\author[1]{Yong Liu\corref{mycorrespondingauthor}}
\ead{yongliu@iipc.zju.edu.cn}

\cortext[mycorrespondingauthor]{Corresponding author}
\fntext[first]{Jun Chen and Shipeng Bai contributed equally to this work.}

\address[1]{Institute of Cyber-Systems and Control, Zhejiang University, China}

\address[address1]{Ningbo Innovation Center, Zhejiang University, China}

\begin{abstract}
Neural network quantization is a very promising solution in the field of model compression, but its resulting accuracy highly depends on a training/fine-tuning process and requires the original data. This not only brings heavy computation and time costs but also is not conducive to privacy and sensitive information protection.
Therefore, a few recent works are starting to focus on data-free quantization.
However, data-free quantization does not perform well while dealing with ultra-low precision quantization.
Although researchers utilize generative methods of synthetic data to address this problem partially, data synthesis needs to take a lot of computation and time.
In this paper, we propose a data-free mixed-precision compensation (DF-MPC) method to recover the performance of an ultra-low precision quantized model without any data and fine-tuning process.
By assuming the quantized error caused by a low-precision quantized layer can be restored via the reconstruction of a high-precision quantized layer, we mathematically formulate the reconstruction loss between the pre-trained full-precision model and its layer-wise mixed-precision quantized model.
Based on our formulation, we theoretically deduce the closed-form solution by minimizing the reconstruction loss of the feature maps.
Since DF-MPC does not require any original/synthetic data, it is a more efficient method to approximate the full-precision model.
Experimentally, our DF-MPC is able to achieve higher accuracy for an ultra-low precision quantized model compared to the recent methods without any data and fine-tuning process.
\end{abstract}

\begin{keyword}
Neural Network Compression, Date-Free Quantization
\end{keyword}

\end{frontmatter}

% \linenumbers

\section{Introduction}

In order to realize the deployment of deep neural networks on resource-constrained lightweight devices, a series of remarkable neural network compression techniques are gradually developing, including low-rank factorization~\cite{yu2017compressing}, parameter and filters pruning~\cite{han2015learning, li2016pruning,wang2021neural,guo2023sensitivity}, quantization~\cite{chen2015compressing, rastegari2016xnor,chen2023learning,chien2023bayesian} and knowledge distillation~\cite{hinton2015distilling,peng2019correlation,cho2023ambiguity,xu2023computation}.
Among these neural network compression techniques, quantization is viewed as a more suitable scheme for hardware acceleration~\cite{han2015deep,chen2020learning} than pruning and knowledge distillation. In this sense, this paper will focus on quantization.

Quantization can be divided into data-driven quantization and data-free quantization~\cite{nagel2019data,cai2020zeroq,zhang2021diversifying,choi2021qimera} according to whether it depends on the data. And data-driven quantization can be further subdivided into quantization-aware training~\cite{han2015deep,zhou2016dorefa,rastegari2016xnor,hubara2017quantized} and post-training quantization~\cite{banner2019post,zhao2019improving,nagel2020up} according to whether it depends on training/fine-tuning.
However, the original training data is not always easily accessible, especially for privacy, security, and deployment in the field.
Therefore, data-free quantization is a vital research direction to achieve a low-precision model without any original data and training.

The accuracy drop of data-free quantization is particularly dramatic when focusing on the ultra-low precision model. Thus, researchers are starting to utilize generative methods~\cite{xu2020generative,cai2020zeroq,liu2021zero}to generate synthetic samples that resemble the distribution of the original dataset and achieve high accuracy. However, generative methods need to cost a lot of computation and time to synthesize data, which conflicts with the concept of data-free.

\begin{figure}[t]
	\centering
	\includegraphics[scale=0.45]{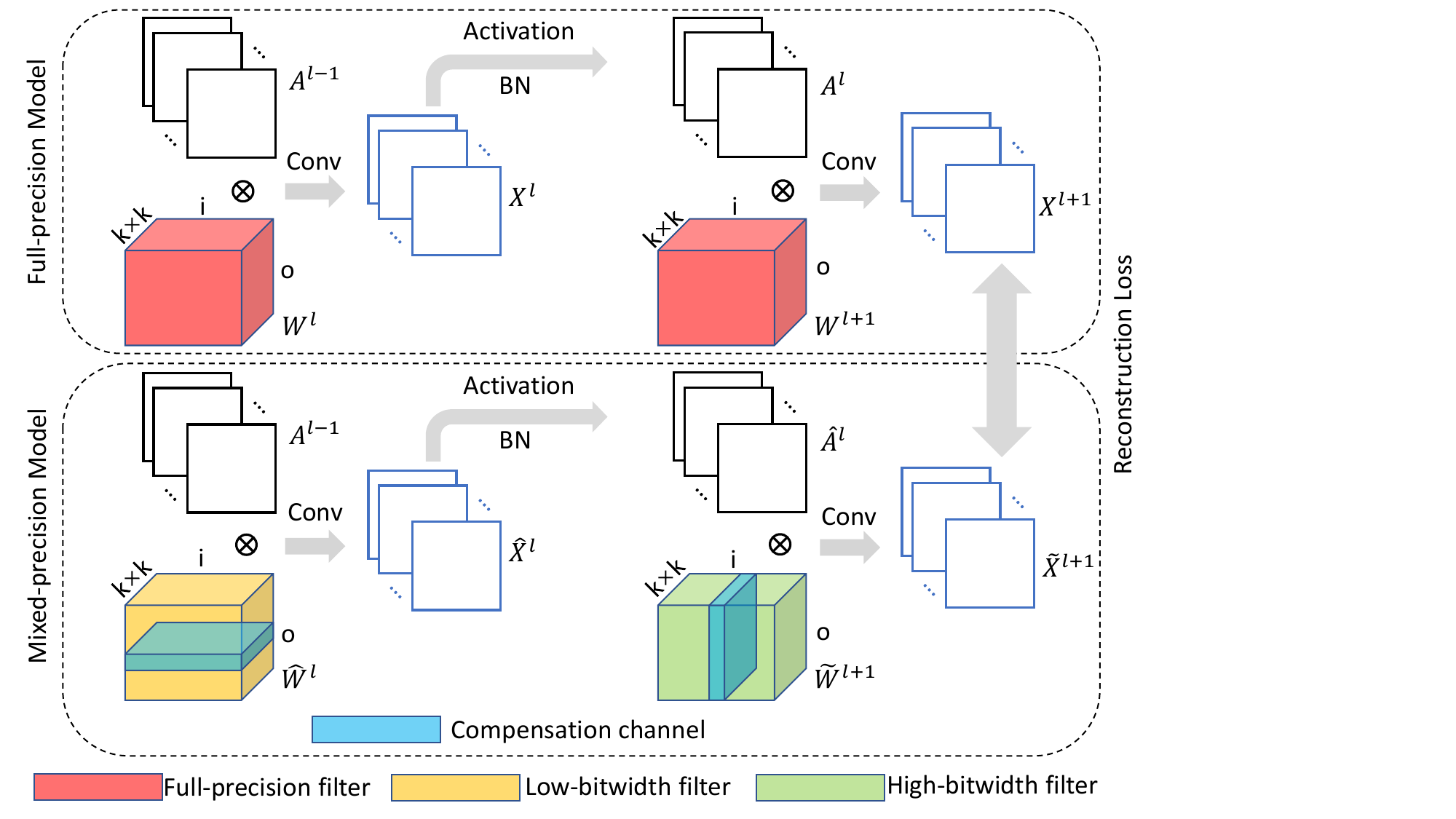}
	\caption{The overview of our DF-MPC method, where the filter in the $l$-th layer is quantized to low-bitwidth and the filter in the $(l+1)$-th layer is quantized to high-bitwidth. The output of $(l+1)$-th convolutional layer can be restored by multiplying the compensation coefficient with respect to the input channel of the high-bitwidth filter, which is equivalent to multiplying the compensation coefficient with respect to the output channel of the low-bitwidth filter. Note that the reconstruction loss is the output difference of $(l+1)$-th layer from the pre-trained full-precision model and its layer-wise mixed-precision quantized model.}
	\label{main}
\end{figure}

In this paper, we abandon the idea of data synthesis and restore the quantized error caused by the ultra-low precision quantization from the perspective of compensation.
Inspired by a few works~\cite{srinivas2015data,kim2020neuron,chu2021mixed}, we propose a data-free mixed-precision compensation (DF-MPC) method to achieve higher accuracy for an ultra-low precision quantization without any data and fine-tuning process, as depicted in Figure~\ref{main}. In summary, we make three main contributions, as shown below:
\begin{itemize}
\item In two adjacent layers of a neural network, we assume that the quantized error caused by a low-precision quantized layer can be restored via the reconstruction of a high-precision quantized layer. Specifically, we quantize the weights in one layer into low precision values (e.g., 2-bit) and then recover the performance by reconstructing relatively higher precision (e.g., 6-bit) weights in the next layer. The layer-wise mixed-precision compensation assumption is described in Section~\ref{sec1}.
\item Based on the mixed-precision compensation assumption, we mathematically formulate the reconstruction loss between the pre-trained full-precision model and its mixed-precision quantized model. Without any fine-tuning process and original/synthetic data, we can achieve layer-wise mixed-precision quantization (e.g., 2/6-bit) only relying on our compensation method.
\item Based on the reconstruction loss, we theoretically deduce the closed-form solution by minimizing the reconstruction loss of the feature maps to restore the quantized error caused by the low precision weight. Furthermore, we verify the effectiveness of our compensation method through experiments on multiple datasets (CIFAR10, CIFAR100, and ImageNet) with multiple network structures (ResNet, DenseNet121, VGG16, and MobileNetV2).
\end{itemize}

\section{Related Work}

Quantization is a kind of model compression method, which accelerates the forward inference phase by converting a full-precision model to a low-precision model (with respect to weights or activations). Whether the low-precision model needs any data or fine-tuning, quantization can continue to be subdivided into the following three classes.

\subsection{Quantization-Aware Training (QAT)}

Since the low-precision representations of weights and activations will cause an accuracy drop, quantization-aware training (QAT) aims to reduce the accuracy drop by retraining or fine-tuning the low-precision with training/validation data~\cite{han2015deep,zhou2016dorefa}. Especially for the ultra-low precision (e.g., binary~\cite{rastegari2016xnor,hubara2017quantized} and ternary~\cite{li2016ternary,zhu2016trained}), QAT can also obtain a satisfactory quantized model.

However, the training process for QAT is computationally expensive and time-consuming. Specifically, the training time and memory of QAT far exceed full precision model training due to simulating quantization operators~\cite{liu2021post}. On the other hand, in some private or secure situations, the original training/validation data is not easy to access.

\subsection{Post-Training Quantization (PTQ)}

Post-training quantization (PTQ) aims to obtain an accurate low-precision model without any fine-tuning process. Therefore, PTQ requires relatively less computation and time consumption than QAT. Specifically, Banner \emph{et al}.~\cite{banner2019post} proposed the 4-bit post-training quantization method that introduces a per-channel allocation and bias-correction, and approximates the optimal clipping value analytically from the distribution of the tensor.
Zhao \emph{et al}.~\cite{zhao2019improving} proposed outlier channel splitting that requires no additional training and works on commodity hardware.
Nagel \emph{et al}.~\cite{nagel2020up} found a good solution to the per-layer weight-rounding mechanism via a continuous relaxation, but this method still requires a small amount of unlabelled data.

Since QAT is fully trained on the entire training data, PTQ's performance tends to be inferior to QAT's regardless of the bit width quantization, which is also the bottleneck of PTQ. And compared to QAT, PTQ is also still not completely free from the original data dependence.

\subsection{Data-Free Quantization (DFQ)}

Compared to QAT and PTQ, data-free quantization (DFQ) requires neither training/validation data nor fine-tuning/training process. In particular, Nagel \emph{et al}.~\cite{nagel2019data} could greatly recover the accuracy of low-precision models by applying weight equalization and bias correction. However, it suffers a huge accuracy drop while the bit width is less than 6-bit. Cai \emph{et al}.~\cite{cai2020zeroq} utilized synthetic data to achieve mixed-precision quantization, but it is also difficult to deal with the accuracy drop below 4-bit.

Recently, researches on DFQ seem to turn to data sampling and generation. Zhang \emph{et al}.~\cite{zhang2021diversifying} proposed a sample generation method that enhances the diversity of data by slacking the alignment of feature statistics in the BN layer and designing a layerwise enhancement. Choi \emph{et al}.~\cite{choi2021qimera} proposed a method that uses superposed latent embeddings to generate synthetic boundary supporting samples, and confirmed that samples near the boundary can improve the performance of a low-precision model. Although DFQ based on data synthesis does not use the original training/validation data, it costs a lot of computation and time to synthesize the data.

\section{Problem Formulation of Data-Free Quantization}

In this section, we present the problem of data-free quantization with the corresponding full-precision pre-trained model.

\subsection{Background and Notations}

 Given a neural network model with $L$ layers, we denote $\mathcal{W}^l \in \mathbb{R}^{o \times i \times k \times k}$ and $\mathcal{A}^{l-1} \in \mathbb{R}^{i \times w \times h}$ as the weight in the $l$-th layer and activation in the $(l-1)$-th layer, where $o$ represents the size of output channels, $i$ represents the size of input channels, $k \times k$ is the size of kernel filters and $w \times h$ is the size of activation maps. Then we obtain the feature maps $\mathcal{X}^l \in \mathbb{R}^{o \times w \times h}$
 \begin{equation}
 \mathcal{X}^l = \mathcal{W}^l \otimes \mathcal{A}^{l-1},
 \end{equation}
 where $\otimes$ is the standard convolution operation.
 By introducing the activation function $f$ and a batch normalization $\operatorname{BN}$, we can finally output the activation map based on the feature map
 \begin{equation}
 \mathcal{A}^l = f\left(\operatorname{BN}\left( \mathcal{X}^l\right)\right).
 \label{noanobn}
 \end{equation}

 Subsequently, we consider the ternary weight tensor in the $l$-th layer that consists of three quantized values $\{-1,0,+1\}$ and a scaling factor $\alpha^l$

 \begin{equation}
 \hat{\mathcal{W}}^l=\left\{\begin{aligned}
 +1, & \text { if } \mathcal{W}^l>\Delta^l \\
 0, & \text { if }\left|\mathcal{W}^l\right| \leq \Delta^l \\
 -1, & \text { if } \mathcal{W}^l<-\Delta^l
 \end{aligned}\right. .
 \label{ternary}
 \end{equation}
 Based on Ternary Weight Networks~\cite{li2016ternary}, we can obtain the optimized layer-wise values of the threshold $\Delta^l$ and the scaling factor $\alpha^l$
\begin{equation}
\begin{aligned}
&\Delta^l=0.7 \mathbb{E}\left(\left|\mathcal{W}^l\right|\right)\\
&\alpha^l=\underset{j \in\left\{j\left|\mathcal{W}^l(j)\right|>\Delta^l\right\}}{\mathbb{E}}\left(\left|\mathcal{W}^l(j)\right|\right).
\end{aligned}
\end{equation}
Since the layer-wise scaling factor $\alpha^l$ can be absorbed into a batch normalization, we can omit $\alpha^l$ and use Eq. (\ref{ternary}) to represent the ternary weight tensor directly.

The new feature map $\hat{\mathcal{X}}^l$ will deviate from the original feature map $\mathcal{X}^l$ when we consider the quantization of the weight tensor, resulting in a rapid accuracy drop of the neural network without fine-tuning, i.e.,
\begin{equation}
\hat{\mathcal{X}}^l = \hat{\mathcal{W}}^l \otimes \mathcal{A}^{l-1} \neq \mathcal{X}^l.
\end{equation}

\subsection{Problem Statement}

\begin{figure}[t]
	\centering
	\includegraphics[trim=150 50 150 80,scale=0.5]{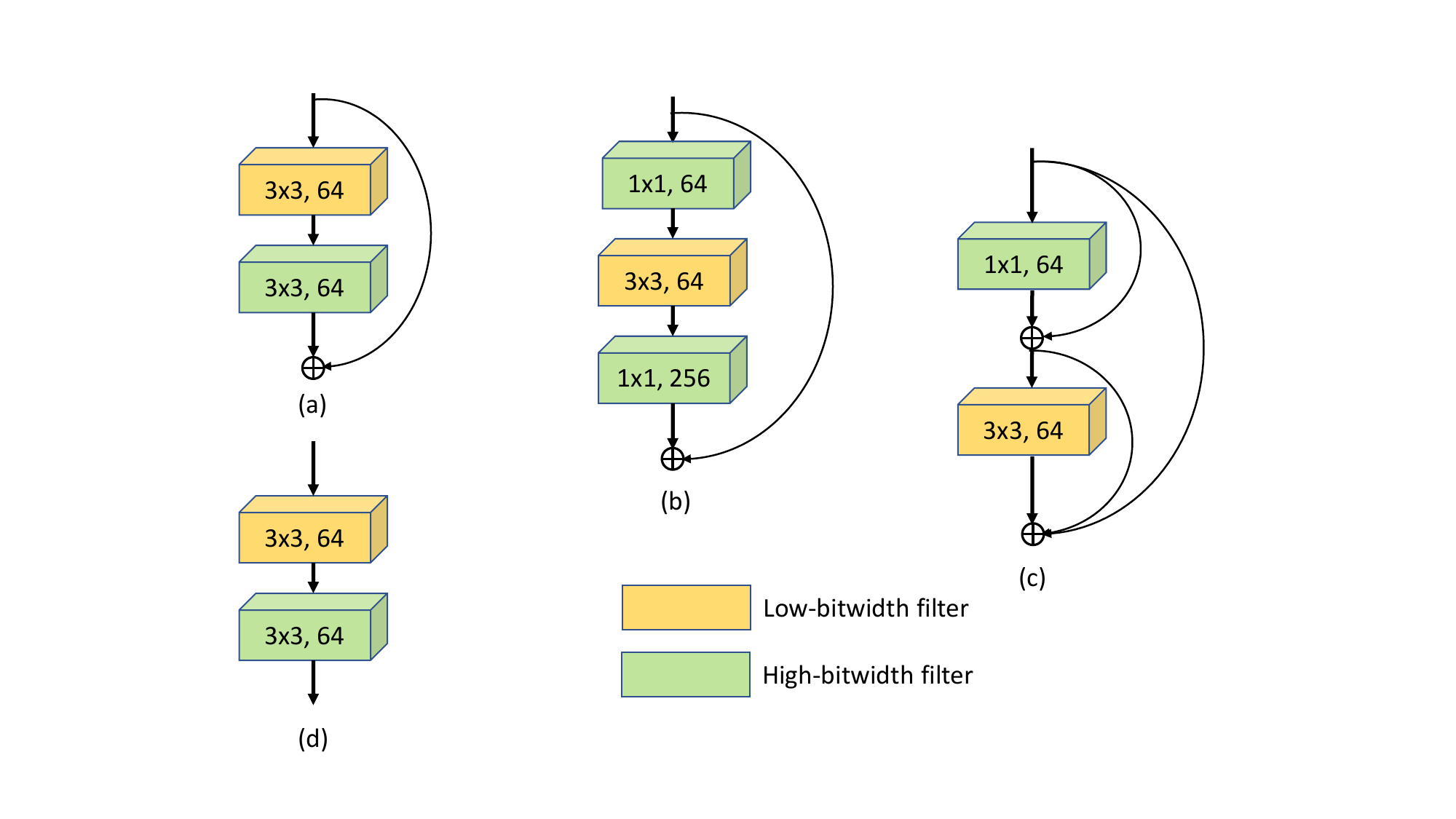}
	\caption{The layer-wise mixed-precision structures of some main deep neural networks. (a): a building block for ResNet18/ResNet34. (b): a bottleneck block for ResNet50/ResNet101. (c): a dense block for DenseNet. (d): a building block for deep neural networks.}
	\label{construction}
\end{figure}

Therefore, we consider reconstructing the weight tensor in the next layer to compensate the feature map in the next layer such that we can recover the accuracy of the low-precision model. Note that we choose a relatively high-precision quantization for the weight tensor of the next layer $\tilde{\mathcal{W}}^{l+1}$ because it is required to compensate the quantized error caused by $\hat{\mathcal{W}}^l$ as much as possible. And we can apply the uniform quantization with $k$-bit based on DoReFa-Net~\cite{zhou2016dorefa}
\begin{equation}
^{k}\mathcal{Q}(\cdot) = \frac{2}{2^k-1} \operatorname{round}\left[(2^k-1)\left(\frac{\cdot}{2\max \vert \cdot\vert} + \frac{1}{2}\right)\right]-1.
\label{uniform}
\end{equation}
Similarly, we omit the layer-wise scaling factor $\max \vert \cdot\vert$ as it can be absorbed into a batch normalization. And we need to make the reconstruction loss between the new feature map and the original feature map as small as possible.

By introducing the coefficient vector $\mathbf{c}=[c_1,c_2,\cdots,c_i]^{T} \geq 0$ whose each component corresponds each input channel of the weight tensor in the $(l+1)$-th layer, we give the $j$-th channel of reconstructed weight tensor as follows:
\begin{equation}
\tilde{\mathcal{W}}^{l+1}_j = c_j \cdot{}^{k}\mathcal{Q}\left( \mathcal{W}_j^{l+1}\right).
\label{reconstruct}
\end{equation}
We hope to find an optimal $\mathbf{c}$ such that the reconstructed feature map $\tilde{\mathcal{X}}^{l+1}_t$ is close to the original feature map $\mathcal{X}^{l+1}_t$, i.e.,
\begin{equation}
\begin{aligned}
&\tilde{\mathcal{X}}^{l+1}_t = \tilde{\mathcal{W}}^{l+1}_{t,j} \otimes \hat{\mathcal{A}}^{l}_j + \sum_{m=1,m\neq j}^{i}\mathcal{W}^{l+1}_{t,m} \otimes \mathcal{A}^{l}_m \\
\approx &\mathcal{X}^{l+1}_t = \sum_{m=1}^{i} \mathcal{W}^{l+1}_{t,m} \otimes \mathcal{A}^l_m.
\end{aligned}
\end{equation}
where the shapes of the weight and activation tensors are $o \times i \times k \times k$ and $i \times w \times h$, respectively. As a result, $\tilde{\mathcal{X}}_t^{l+1}$ and $\mathcal{X}_t^{l+1}$ indicate the $t$-th output channel of the feature map. Note that when we use the same notation $j$ or $m$ to indicate the channel of the weight and activation, it means that their dimensions are the same and correspond to each other in the computation.

Consequently, our problem aims to find a coefficient vector $\mathbf{c}$ to minimize the reconstruction loss based on a full-precision pre-trained model $\mathcal{W}$ without any training process and data, i.e.,
\begin{equation}
\min_{\mathbf{c}}\sum_{t=1}^{o} \lVert \tilde{\mathcal{X}}_t^{l+1} - \mathcal{X}_t^{l+1} \rVert_{2}^2,
\label{loss}
\end{equation}
Note that we apply the mixed-precision quantization, i.e., one layer low-bitwidth (ternary) and one layer high-bitwidth that is used for compensation. The mixed-precision structures of some main deep neural networks are shown in Figure~\ref{construction}.

Although we consider restoring the quantized error for the ternary values, our method is not limited to the ternary case, but is also applicable to higher precision case (even the same as the precision of the quantized filter). For example, we have different mixed-precisions, such as 2/6-bit, 3/6-bit, 6/6-bit etc. Note that in this paper, we use the ternary filter just to distinguish it from the quantized filter.

\section{Proposed Method of Mixed-Precision Compensation}

In this section, we theoretically give the layer-wise mixed-precision compensation assumption for the reconstruction loss of Eq. (\ref{loss}). According to this assumption, we present our data-free mixed-precision compensation method to recover the accuracy of the low-precision neural network.

\subsection{Compensation Assumption}
\label{sec1}

In order to minimize Eq. (\ref{loss}) without any data and fine-tuning process, we assume that the quantized error of each filter with low-bitwidth can be partly compensated by reconstructing filters with high-bitwidth in the next layer.
Then we further assume that the reconstructed filter consists of a linear combination of the high-bitwidth filters and the coefficient value, which is defined as Eq. (\ref{reconstruct}).

\begin{assumption}
	In order to minimize Eq. (\ref{loss}) with a data-free version, we propose a one-to-one channel-wise compensation assumption that the quantized error caused by the low-bitwidth quantization of each channel of the filter can be compensated by the high-bitwidth quantization of the corresponding channel of the filter in the next layer.
\end{assumption}

Without loss of generality, let the filter of the $l$-th layer be quantized to low-bitwidth (ternary) such that the $t$-th channel of the reconstruction loss in the $(l+1)$-th layer can be represented as
\begin{equation}
\begin{aligned}
&\tilde{\mathcal{X}}^{l+1}_t - \mathcal{X}^{l+1}_t = \tilde{\mathcal{W}}^{l+1}_{t,j} \otimes \hat{\mathcal{A}}^{l}_j - \mathcal{W}^{l+1}_{t,j} \otimes \mathcal{A}^{l}_j \\
=& c_j \cdot{}^{k}\mathcal{Q}\left(\mathcal{W}^{l+1}_{t,j}\right) \otimes \hat{\mathcal{A}}^{l}_j - \mathcal{W}^{l+1}_{t,j} \otimes \mathcal{A}^{l}_j \\
=& c_j\cdot\left({}^{k}\mathcal{Q}\left(\mathcal{W}^{l+1}_{t,j}\right) - \mathcal{W}^{l+1}_{t,j}\right) \otimes \hat{\mathcal{A}}^{l}_j + c_j\cdot\mathcal{W}^{l+1}_{t,j} \otimes \hat{\mathcal{A}}^{l}_j - \mathcal{W}^{l+1}_{t,j} \otimes \mathcal{A}^{l}_j \\
=& c_j\cdot\left({}^{k}\mathcal{Q}\left(\mathcal{W}^{l+1}_{t,j}\right) - \mathcal{W}^{l+1}_{t,j}\right) \otimes \hat{\mathcal{A}}^{l}_j + \mathcal{W}^{l+1}_{t,j} \otimes (c_j \cdot\hat{\mathcal{A}}^{l}_j - \mathcal{A}^{l}_j).
\end{aligned}
\label{reconstruction}
\end{equation}
Note that the $l$-th output channel size of $\mathcal{A}$ and $\mathcal{W}$ is equal to the $(l+1)$-th input channel size of $\mathcal{W}$.
For brevity, we first omit the activation function $f$ and a batch normalization $\operatorname{BN}$. Then the equations $\hat{\mathcal{A}}^l_j=\hat{\mathcal{X}}^l_j$ and $\mathcal{A}^l_j=\mathcal{X}^l_j$ hold. By introducing the two formulas
\begin{equation}
\begin{aligned}
&\hat{\mathcal{A}}^l_j=\hat{\mathcal{X}}^l_j = \sum_{m=1}^{i} \hat{\mathcal{W}}^{l}_{j,m} \otimes \mathcal{A}^{l-1}_m \\
&\mathcal{A}^l_j=\mathcal{X}^l_j = \sum_{m=1}^{i} \mathcal{W}^{l}_{j,m} \otimes \mathcal{A}^{l-1}_m.
\end{aligned}
\label{convolution}
\end{equation}

\section{Experiments}

\begin{figure}[t]
	\centering
	\includegraphics[scale=0.4]{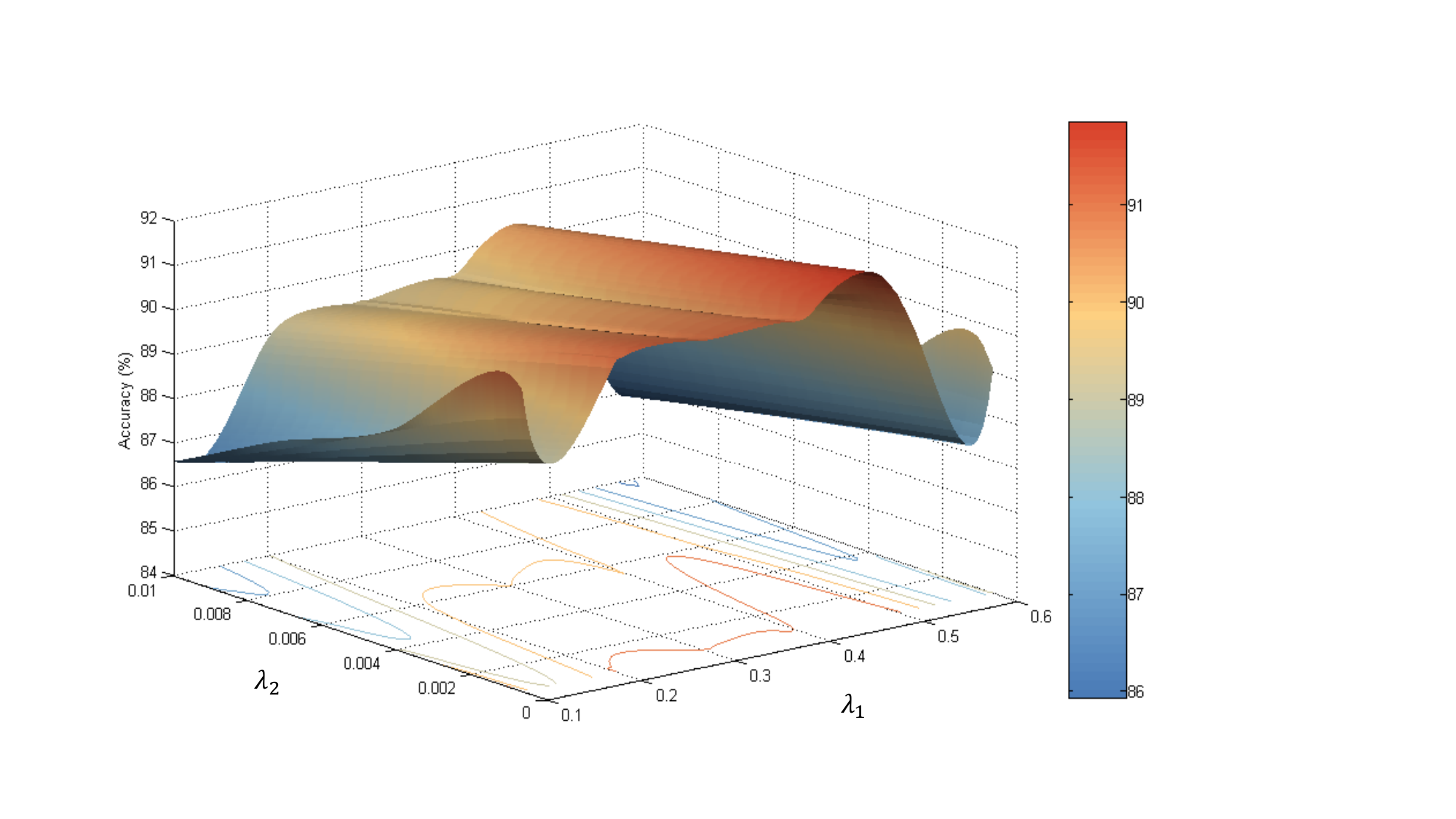}
	\caption{The accuracy comparison of different $\lambda_1$ and $\lambda_2$ values in Eq. (\ref{solution}). On CIFAR10 with ResNet56, $\lambda_1$ and $\lambda_2$ vary from 0.1 to 0.6 and from 0 to 0.01, respectively.}
	\label{ablation}
\end{figure}

In this section, we evaluate our method on CIFAR10/CIFAR100~\cite{krizhevsky2009learning} and ImageNet~\cite{krizhevsky2017imagenet} datasets, which are well-known datasets for evaluating the performance on the image classification.

\textbf{Dataset.} CIFAR10/CIFAR100 datasets consist of 50k training sets and 10k validation sets, which are natural color images with 32$\times$32 for small-scale experiments. CIFAR10 dataset is organized into 10 classes and CIFAR100 dataset into 100 classes, respectively.
ImageNet dataset consists of 1.2 million training sets and 50k validation sets, which are high-resolution natural images for large-scale experiments. These images are organized into 1000 categories.

\begin{table}[t]
	\caption{Top-1 classification accuracy results on CIFAR10 dataset with ResNet18, ResNet56, and VGG16. FP32 denotes the full-precision weights. MP2/6 denotes the layer-wise 2 bit and 6 bit mixed-precision weights.}
	\begin{center}
		\begin{tabular}{cccc}
			\toprule
			Model & Method & FP32 (\%) & MP2/6 (\%) \\
			\midrule
			\multirow{2}{*}{ResNet18} & Original & 92.61 & 10.78  \\
			& DF-MPC & 92.61 & 89.12  \\ \midrule
			\multirow{2}{*}{ResNet56} & Original & 93.88 & 38.03  \\
			& DF-MPC & 93.88 & 91.05  \\ \midrule
			\multirow{2}{*}{VGG16} & Original & 93.70 &  10.00 \\
			& DF-MPC & 93.70 & 90.48 \\
			\bottomrule
		\end{tabular}
	\end{center}
	\label{table1}
\end{table}
\begin{table}[H]
	\caption{Top-1 classification accuracy results on CIFAR100 dataset with ResNet18 and VGG16. FP32 denotes the full-precision weights. MP2/6 denotes the layer-wise 2 bit and 6 bit mixed-precision weights.}
	\begin{center}
		\begin{tabular}{cccc}
			\toprule
			Model & Method & FP32 (\%) & MP2/6 (\%) \\
			\midrule
			\multirow{2}{*}{ResNet18} & Original & 73.62 & 1.05  \\
			& DF-MPC & 73.62 & 64.90  \\ \midrule
			\multirow{2}{*}{VGG16} & Original & 70.09 &  3.80 \\
			& DF-MPC & 70.09 & 64.95 \\
			\bottomrule
		\end{tabular}
	\end{center}
	\label{table2}
\end{table}

\textbf{Model.} We choose ResNet~\cite{he2016deep} (including ResNet18, ResNet50, ResNet56, ResNet101), DenseNet121~\cite{huang2017densely}, VGG16~\cite{simonyan2014very} and MobileNetV2~\cite{sandler2018mobilenetv2} for evaluation. All the model and pre-trained full-precision weights are from pytorchcv library~\url{https://pypi.org/project/pytorchcv/}.

\textbf{Setting.} We implement our method using PyTorch~\cite{paszke2017automatic} and run the experiments using GTX 1080Ti.

\subsection{Ablation Study on CIFAR}

\begin{table}[t]
	\caption{Top-1 classification accuracy results on ImageNet dataset with ResNet.}
	\begin{center}
		\begin{tabular}{ccccc}
			\toprule
			Model & Method & W-bit & Size (MB) & Top-1 Acc (\%) \\
			\midrule
			\multirow{5}{*}{ResNet18} & Full-precision & 32 & 44.59 & 71.47 \\
			& OMSE~\cite{choukroun2019low} & 4 & 5.58 & 64.03 \\
			& GZNQ~\cite{he2021generative} & 4 & 5.58 & 64.50 \\
			& DFQ~\cite{nagel2019data} & 6 & 8.36 & 66.30 \\
			& DF-MPC & 2/6 & \textbf{5.48} & \textbf{66.46} \\ \midrule
			\multirow{4}{*}{ResNet50} & Full-precision & 32 & 97.49 & 76.12 \\
			& OCS~\cite{zhao2019improving} & 4 & 12.28 & 69.30 \\
			& OMSE~\cite{choukroun2019low} & 4 & 12.28 & 70.06 \\
			& DF-MPC & 2/6 & \textbf{10.55} & \textbf{71.20} \\ \midrule
			\multirow{3}{*}{ResNet101} & Full-precision & 32 & 170.41 & 77.31 \\
			& OMSE~\cite{choukroun2019low} & 4 & 21.30 &  71.49 \\
			& DF-MPC & 2/6 & \textbf{18.36} & \textbf{72.59} \\
			\bottomrule
		\end{tabular}
	\end{center}
	\label{table3}
\end{table}

We first conduct a series of ablation studies on CIFAR datasets to investigate the effect of components of the proposed DF-MPC scheme. We evaluate our method on MP2/6 weights and FP32 activations.

Based on Eq. (\ref{solution}), our method has two regularization coefficients $\lambda_1$ and $\lambda_2$ that affect the effect of compensation directly.
Specifically, we adjust these two hyper-parameters to find the optimal solution, as shown in Figure~\ref{ablation}.
On the one hand, as $\lambda_1$ varies from 0.1 to 0.5, the final accuracy of the quantized model increase steadily. But it suffers a significant drop when $\lambda_1$ is set to 0.6.
On the other hand, the final performance is mainly on the decline when $\lambda_2$ varies from 0 to 0.01.
In summary, the compensation combination of $\lambda_1=0.5$ and $\lambda_2=0$ is the optimal solution for ResNet56 on CIFAR10 dataset.

For $\lambda_2=0$, we also verify from this ablation study that the constraint $\Vert \mathbf{c}\Vert^2_{2}$ in Eq. (\ref{4loss}) does not work, which is consistent with our theoretical analysis, i.e., the term $\left({}^{k}\mathcal{Q}\left(\mathcal{W}^{l+1}_{t,j}\right) - \mathcal{W}^{l+1}_{t,j}\right)$ has very little effect.
For $\lambda_1=0.5$, we know that in order of importance, $\Vert \Gamma\Vert^2_{2}$ is greater than $\Vert \Theta\Vert^2_{2}$.

Table~\ref{table1} and Table~\ref{table2} show the performance before and after compensation on CIFAR10 and CIFAR100 datasets, respectively.
If the full-precision model is quantized to a mixed-precision of 2-bit and 6-bit directly, its accuracy will become no different from random initialization.
However, after our compensation method, the same quantization mode will result in a fully usable quantized model with great accuracy improvement. Experimentally, this also proves the effectiveness of our DF-MPC.

\subsection{Experiments on ImageNet}

\begin{table}[h]
	\caption{Top-1 classification accuracy results on ImageNet dataset with DenseNet121 and MobileNetV2.}
	\begin{center}
		\begin{tabular}{ccccc}
			\toprule
			Model & Method & W-bit & Size (MB) & Top-1 Acc (\%) \\
			\midrule
			\multirow{4}{*}{DenseNet121} & Full-precision & 32 & 31.92 & 74.36 \\
			& OCS~\cite{zhao2019improving} & 4 & 4.09 & 63.00 \\
			& OMSE~\cite{choukroun2019low} & 4 & 4.09 & 64.40 \\
			& DF-MPC & 3/6 & \textbf{3.39} & \textbf{70.02} \\ \midrule
			\multirow{5}{*}{MobileNetV2} & Full-precision & 32 & 13.37 & 73.03 \\
			& GDFQ~\cite{xu2020generative} & 6 & 2.50 & 70.98 \\
			& GZNQ~\cite{he2021generative} & 6 & 2.50 & 71.12 \\
			& DFQ~\cite{nagel2019data} & 8 & 3.34 & 71.20 \\
			& DF-MPC & 6 & \textbf{2.50} & \textbf{71.29} \\
			\bottomrule
		\end{tabular}
	\end{center}
	\label{table4}
\end{table}

We evaluate our method on ImageNet dataset for the large-scale image classification task, and compare the performance with other data-free quantization methods over various models. Here, GDFQ~\cite{xu2020generative} and GZNQ~\cite{he2021generative} are the generative methods and they still utilize synthetic data to complete the quantization.

Table~\ref{table3} and Table~\ref{table4} compare the performance with previous methods, such as OCS~\cite{zhao2019improving}, DFQ~\cite{nagel2019data}, and OMSE~\cite{choukroun2019low}. For 2-bit, our DF-MPC uses the ternary representation based on Eq. (\ref{ternary}). For 3-bit and 6-bit, our DF-MPC uses the quantized representation based on Eq. (\ref{uniform}). Based on layer-wise mixed-precision compensation, we achieve higher accuracy at the smaller model size.
In particular, our method with 3/6-bit outperforms DFQ~\cite{nagel2019data} with 6-bit by 0.16\% on ResNet18. And our method with 6-bit outperforms DFQ~\cite{nagel2019data} with 8-bit by 0.09\% on MobileNetV2.
Note that our 6-bit scheme actually implies 6/6-bit mixed-precision quantization.

\begin{figure}[t]
	\centering
	\subfigure[ResNet56 in layer1-2]
	{\includegraphics[trim=150 300 150 350,scale=0.66]{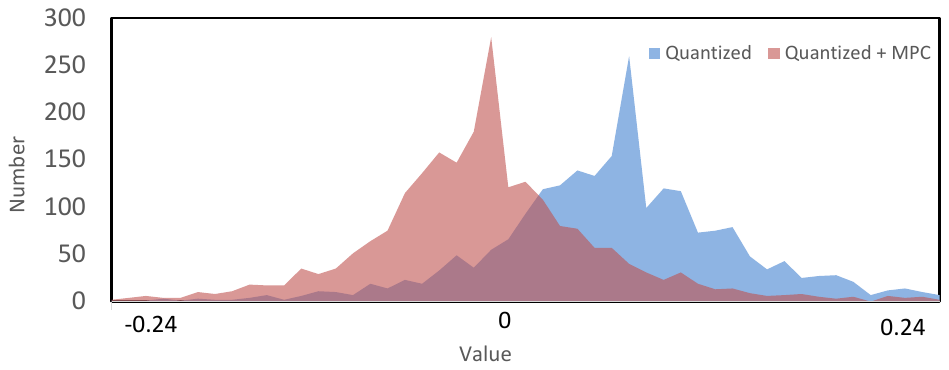}}
	\subfigure[ResNet56 in layer3-8]
	{\includegraphics[trim=150 300 150 350,scale=0.65]{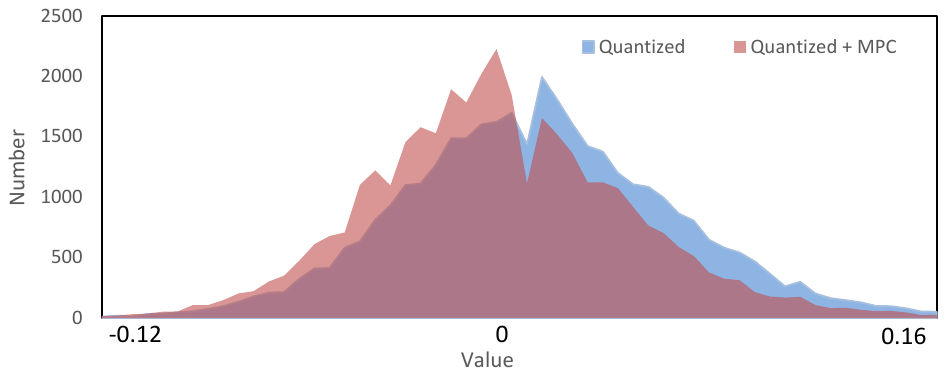}}
	\caption{The 6-bit quantized weight distribution before and after compensation on CIFAR10 dataset. The mean of the compensated weight distribution is closer to zero.}
	\label{distribution}
\end{figure}

\textbf{DF-MPC} \emph{vs.} \textbf{ZeroQ.} The generative methods need to cost a lot of computation and time due to data synthesis. For example, ZeroQ~\cite{cai2020zeroq} of ResNet18 takes 12 seconds on an 8-V100 system.
In contrast, DF-MPC of ResNet18 takes only 2 seconds on a single GTX 1080 Ti, or can even run on CPU only, which makes the deployment of quantized models convenient and fast.

\textbf{DF-MPC} \emph{vs.} \textbf{DFQ.} DFQ~\cite{nagel2019data} and DF-MPC have some common ideas. DFQ also considers the relation between the output channel in the $l$-th layer and the input channel in the $(l+1)$-th layer. Specifically, DFQ scales the cross-layer factor to equalize the weight tensor channel ranges.
However, DF-MPC scales the cross-layer factor to minimize the output difference of feature maps in the $(l+1)$-th layer between the pre-trained full-precision model and its layer-wise mixed-precision quantized model.
Theoretically, our method guarantees the minimal quantized error of the layer-wise mixed-precision model.

\subsection{Visualization}

\begin{figure}[t]
	\centering
	\subfigure[Before Compensation]
	{\includegraphics[width=0.49\textwidth]{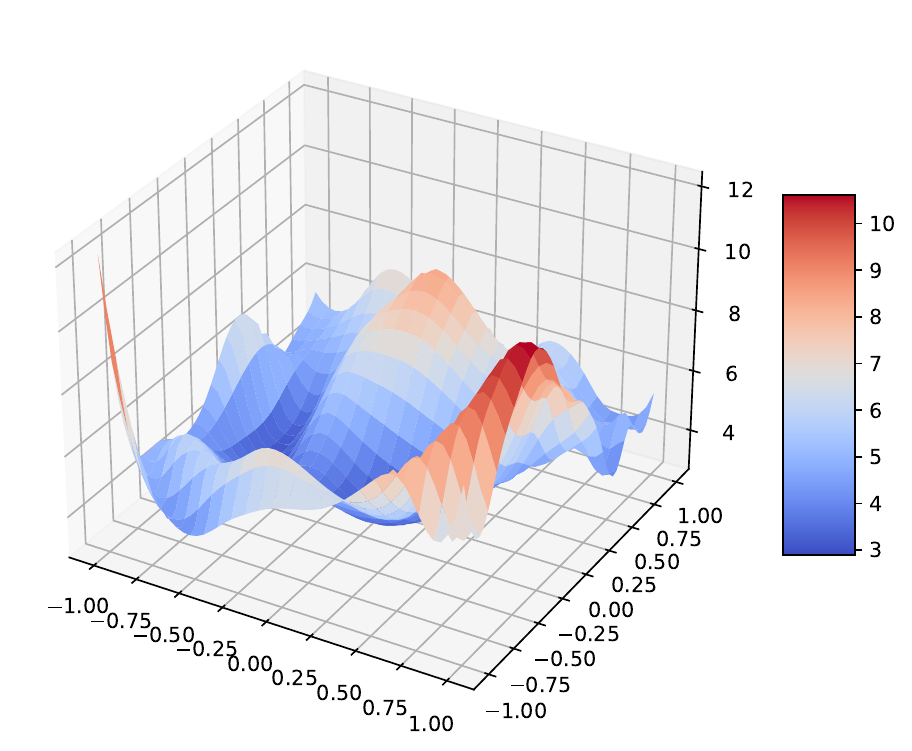}}
	\subfigure[After Compensation]
	{\includegraphics[width=0.49\textwidth]{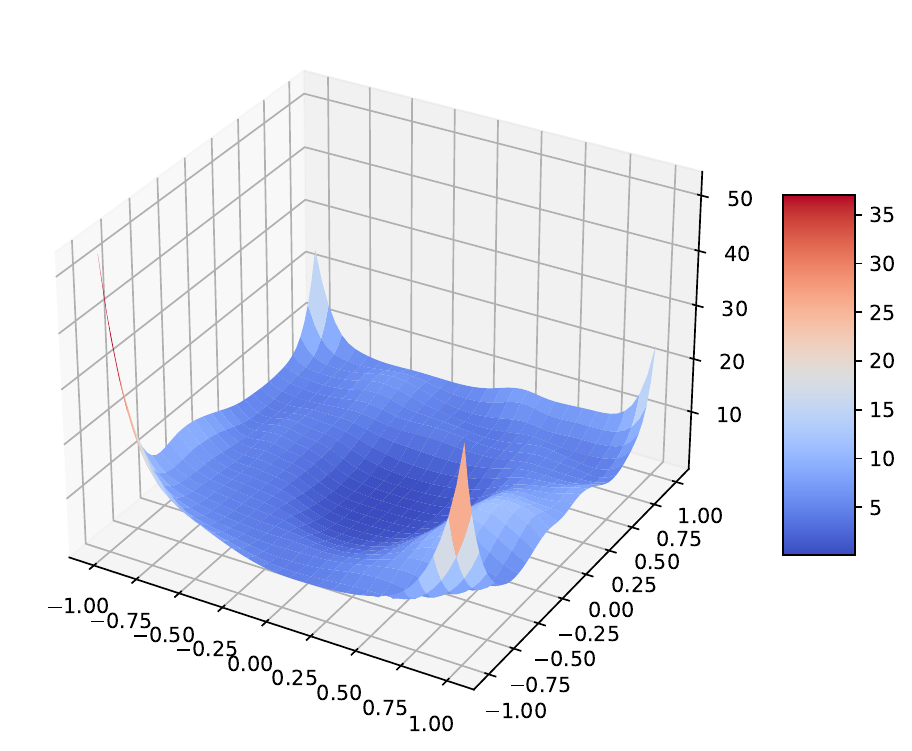}}
	\caption{The loss surfaces of the mixed-precision ResNet56 before and after compensation on CIFAR10 dataset, which reflects the sharpness/flatness of different quantized weights.}
	\label{losssurface}
\end{figure}

Figure~\ref{distribution} shows the quantized weight distribution before and after compensation in two different layers of ResNet18. After our DF-MPC method, the mean of the 6-bit quantized weight distribution approaches zero.
Moreover, based on the previous work~\cite{li2018visualizing}, we show the loss surfaces before and after compensation. By analyzing Figure~\ref{losssurface}, we find that the loss landscape of the quantized model before compensation is sharp, which shows no noticeable convexity.
On the contrary, the loss landscape of the quantized model after compensation is smooth and flat, and shows noticeable convexity, which is consistent with the pre-trained full-precision model.

\section{Conclusion}

This paper proposed the problem of recovering the accuracy of an ultra-low precision model without any data and fine-tuning, which only relies on the pre-trained full-precision model. By assuming the quantized error caused by a low-precision quantized layer can be restored via the reconstruction of a high-precision quantized layer, we mathematically formulated the reconstruction loss of the feature maps between the pre-trained full-precision model and its mixed-precision quantized model. Based on our formulation, we designed a data-free mixed-precision compensation method along with its closed-form solution.

Since no original/synthetic data is used, we can not access the feature maps, which leads to our method being slightly worse than generative methods with synthetic data.
Our future work would extend an expert neural network to estimate the feature maps in the reconstruction loss, which further recovers the performance of the quantized model.

\bibliography{mybibfile}

\end{document}